\documentclass{article}
\usepackage{ijcai21}

\usepackage{times}

\usepackage{soul}
\usepackage{url}
\usepackage[hidelinks]{hyperref}
\usepackage[utf8]{inputenc}
\usepackage[small]{caption}
\usepackage{graphicx}
\usepackage{amsmath}
\usepackage{booktabs}
\usepackage{times}
\usepackage{epsfig}
\usepackage{graphicx}
\usepackage{amsmath}
\usepackage{amssymb}
\usepackage{mathtools}
\usepackage{soul}
\usepackage[dvipsnames]{xcolor}
\newcommand{\comment}[1]{}
\newcommand\mydef{\mathrel{\stackrel{\makebox[0pt]{\mbox{\normalfont\tiny def}}}{=}}}
\newcommand{\lc}[1]{\textcolor{black}{#1}}
\newcommand{\ca}[1]{\textcolor{Apricot}{#1}}
\newcommand{\cb}[1]{\textcolor{Blue}{#1}}
\title{SPeCiaL: Self-Supervised Pretraining for Continual Learning}

\author{
Lucas Caccia$^{1,2}$\footnote{Contact Author}\and
Joelle Pineau$^{1,2}$\\
\affiliations
$^1$McGill University\\
$^2$Facebook AI Research
\emails
lucas.page-caccia@mail.mcgill.ca
}

\begin{document}

\maketitle

\begin{abstract}

This paper presents SPeCiaL: a method for unsupervised pretraining of representations tailored for continual learning. Our approach devises a meta-learning objective that differentiates through a sequential learning process. Specifically, we train a linear model over the representations to match different augmented views of the same image together, each view presented sequentially. The linear model is then evaluated on both its ability to classify images it just saw, and also on images from previous iterations. This gives rise to representations that favor quick knowledge retention with minimal forgetting. We evaluate SPeCiaL in the Continual Few-Shot Learning setting, and show that it can match or outperform other supervised pretraining approaches.


\end{abstract}

\section{Introduction}

\noindent
In Machine Learning, practitioners often rely on the ubiquitous assumption that the training data is sampled independently and identically from a fixed distribution. This iid assumption is a convenient requirement, spanning from classical approaches to Deep Learning based methods. Yet, the world that humans encounter is far from static: the data we observe is temporally correlated, and changes constantly. Therefore many practical applications, from autonomous driving to conversational chatbots, can greatly benefit from the ability to learn from non-stationary data. 

\noindent
To this end, the field of Continual Learning\cite{french1991using,kirkpatrick2017overcoming} tackles the problem setting where an agent faces an online and non-stationary data stream. The core of CL methods have been intently focused on directly minimizing catastrophic interference, a phenomenon whereby previously acquired knowledge gets overwritten as the agent learns new skills. 



In this paper, we are interested in the deployment of CL systems \cite{osaka}. Namely, we are interested in \emph{preparing} a system before its deployment on non-stationary data where it will need to make predictions and accumulate knowledge in an online fashion. 
We believe this setting captures better the deployment of CL systems in real life, as it would be more realistic to deploy an agent with some - albeit limited - knowledge of the world. 

Therefore, we ask ourselves how can we equip agents with data representations that are amenable to learning under non-stationarity? An innovative solution was first proposed by \cite{javed2019meta}, coined \textit{Online-Aware Meta Learning} (OML). 
\lc{ This objective differentiates through a supervised sequential learning task and optimizes for minimal forgetting. }
While ingenious, OML requires access to labelled data, which can be difficult or expensive to obtain for many CL scenarios. For example, is it fairly cheap to obtain large unlabelled datasets from a dashboard camera in the hopes of training a self-driving agent. However, manually annotating each frame - or even deciding which types of annotation - can be time consuming and costly.  


In this work, we instead aim to learn representations amenable to CL purely from unlabelled data. Recently, the field of unsupervised representation learning (or self-supervised learning) has shown tremendous progress \cite{chen2020simple}, closing the gap with fully supervised methods on large-scale image datasets. Moreover, SSL methods have shown better transfer capabilities \cite{chen2020simple} to out-of-distribution data than supervised methods. To this end, we leverage key components of recent SSL methods, enabling us to devise an efficient label-free representation learning algorithm for CL. Concretely, our contributions are as follows : 

\begin{itemize} 
\item We propose a novel strategy, called SPeCiaL, for unsupervised pretraining of representations amenable to continual learning
\item Our algorithm efficiently reuses past computations, reducing its computational and memory footprint.
\item We show that SPeCiaL can match or outperform OML on several continual few-shot learning benchmarks. 

\end{itemize}

\section{Related Work}
Our work builds upon several research fields, which we summarize here. 
\noindent

\paragraph{Self-Supervised Learning} The field of Unsupervised Learning of Visual Representations has made significant progress over the last few years, reducing the gap with supervised pretraining on many downstream tasks. By using data augmentations, SSL methods can generate pseudo-labels, matching different augmented views of the same image together. Instance-level classifiers \cite{chen2020simple,chen2020improved} treat each image - and its augmented versions -  as its own class, and rely on noise contrastive estimation to compare latent representations. Clustering Methods \cite{caron2020unsupervised} instead group latent instances together, either using the cluster id as a label, or ensuring cluster consistency across different views of the same image. \lc{In general, SSL methods are not designed to be deployed in non-stationary settings, which can lead to suboptimal performance (see section \ref{experiments}). Our proposed method addresses this specific issue. } 

\paragraph{Continual Learning}
The supervised CL literature is often divided into several families of methods, all with the principal objective of reducing catastrophic forgetting. In the fixed architecture setting we have \textit{prior based methods}~\cite{kirkpatrick2017overcoming,zenke2017continual,nguyen2017variational}, which regularize the model to limit the change of parameters deemed important in previous tasks. \textit{Rehearsal based methods} ~\cite{chaudhry2019continual,aljundi2019online}, fight forgetting by replaying past samples alongside the incoming data. This data can be store in a small episodic memory \cite{aljundi2019gradient} or sampled from a generative model also trained continually ~\cite{shin2017continual,van2018generative}. Some replay based methods further aim at directly optimizing for the reduction of negative interference across tasks  \cite{lopez2017gradient,riemer2018learning,chaudhry2018efficient,NEURIPS2020_85b9a5ac}.

\noindent
\textbf{Unsupervised Continual Learning} Several works \cite{rao2019continual,smith2019unsupervised,achille2018life} learn representations in an unsupervised manner over non-iid data. We also note that some generative replay methods also fit into this category, namely ones where the generator does not leverage labelled data \cite{caccia2020online,shin2017continual}. This line of work differs from the setting explored in this paper, as the representation learning occurs not during pretraining (where we assume iid acess to some data) but over a stream of non-stationary data.


\noindent
\paragraph{Meta-Learning} The Meta-Learning \cite{ravi2016optimization,snell2017prototypical,vinyals2016matching} - or learning to learn -  field aims to design models which can acquire new skills or knowledge fast, via a small set of examples. Most of the literature has focused on the few-shot supervised learning setting, where agents are first pretrained offline and evaluated on their ability to label new classes given small support sets. Model-Agnostic Meta-Learning (MAML)\cite{finn2017model} enables agents to learn a parameter initialization suitable for fast adaption, by directly optimizing through the update procedure performed by the model at deployment time. The flexibility of MAML has led its use to reach beyond few-shot learning. 

\paragraph{Meta-Continual Learning} Most relevant to our work is Online-Aware Meta-Learning (OML)\cite{javed2019meta}, which designs a meta-objective using catastrophic interference as a training signal. In the pretraining step, the OML objective artificially creates a temporally correlated sequence through which the agent must learn. Follow up work proposed ANML \cite{beaulieu2020learning}, a neuromodulated architecture which meta-learns a gating mechanism to reduce forgetting. Differently, recent works have propose the use of meta-learing objectives at continual-learning time \cite{NEURIPS2020_85b9a5ac,osaka}.
\paragraph{Unsupervised Meta-Learning} Variants of MAML which do not require labelled data in the pretraining phase have been proposed \cite{hsu2018unsupervised,antoniou2019assume,khodadadeh2018unsupervised}. \lc{These methods however are not designed for continual learning, as they have no mechanism to prevent catastrophic forgetting.}
Lastly and closest to our work is \cite{bertugli2021generalising}, where the authors propose a unsupervised pretraining strategy for learning representations in the continual few shot learning setting. Similar to \cite{hsu2018unsupervised}, the authors use an unsupervised embedding function with clustering to label examples, which are then used within the OML algorithm, resulting in a 3-step procedure for the pretraining phase. \lc{Our work instead proposes a simplified (single step) pretraining procedure, which is both conceptually simpler, and is easier to extend.}


\begin{figure*}[ht]
    \begin{minipage}{0.95\textwidth}
    \includegraphics[width=1.0\textwidth]{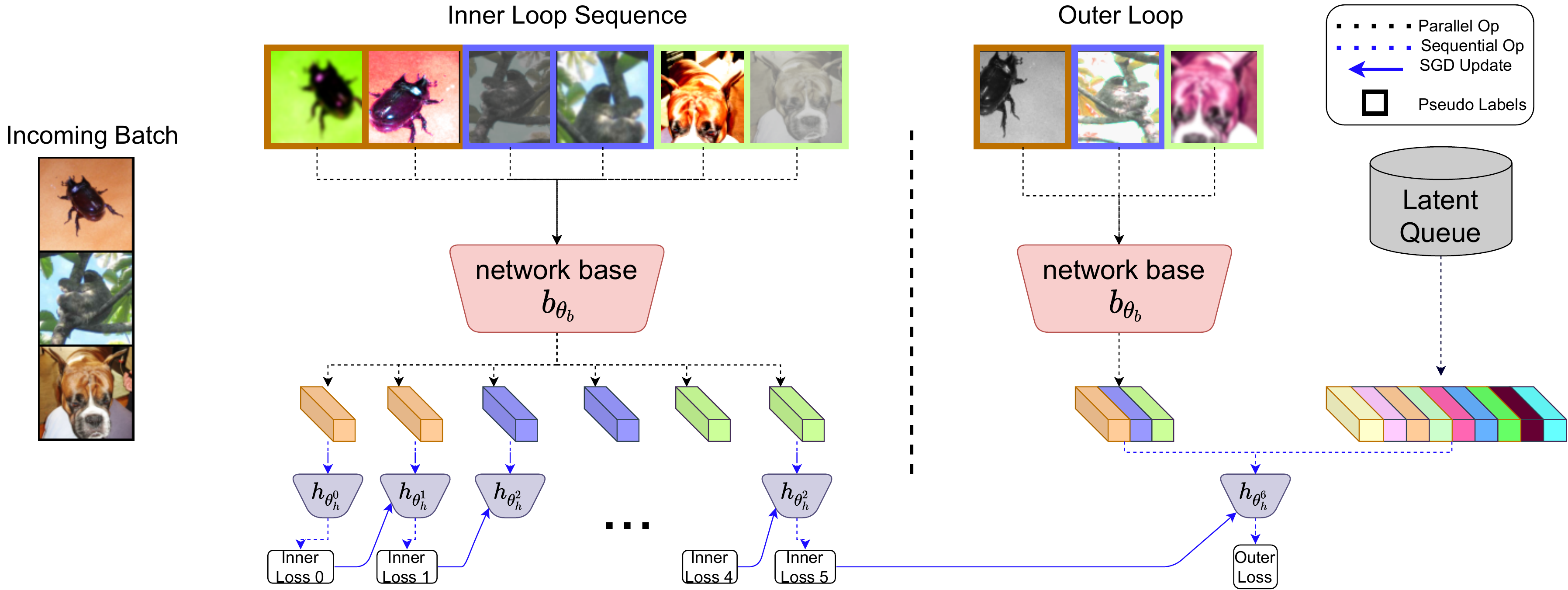}
    \end{minipage}\hfill
    \vspace{10pt}
    \caption{Workflow of our proposed method, SPeCiaL. From the incoming batch we create a temporally correlated sequence with augmentations from the same image shown next to each other. Augmentations of the same image are assigned a unique label, which the model learns during the inner loop. In the outer loop the model is evaluated both on its retention of the seen sequence, but also on past samples seen in previous updates. The encoder $f_{\theta}$ extracts all representations in parallel, leaving only the linear head $h_{\phi}$ to run sequentially.}
    \label{fig:cartoon}
\end{figure*}

\section{Background: Meta Continual Learning}

In this section we formalize the learning setting under which our proposed method operates. We begin by first defining the general continual learning setting, in which (potentially pretrained) agents are deployed. We then describe meta-continual learning, a framework to pretrain agents before deployment in a continual learning setting.

We follow \cite{lopez2017gradient,javed2019meta} and define a Continual Learning Problem (CLP) as a potentially infinite stream of samples:
$$
\mathcal{S} = (X_1, Y_1), (X_2, Y_2), \hdots, (X_t, Y_t), \hdots,  
$$
\noindent
where input $X_t$ and output $Y_t$ belong to sets $\mathcal{X}$ and $\mathcal{Y}$ respectively. Crucially, examples are not drawn i.i.d from a fixed distribution over $(X,Y)$ pairs. Instead, we assume that at each timestep $t$, input samples are drawn according to the conditional distribution $P(X_t|C_t)$, where $C_t \in \mathcal{C}$ is a hidden \textit{time-dependent context variable}  \cite{osaka}, denoting the state of the distribution at time $t$. In practice, $C_t$ can be the previous frames of a video, or the cards currently dealt in a poker game. Given this context variable, samples can be drawn i.i.d. from $P(X_t|C_t)$, in other words the $C_t$ encapsulates all the non-stationarity of the CLP. 

Our goal is to learn a $\theta$-parameterized predictor $f_{\theta} : \mathcal{X} \rightarrow \mathcal{Y}$ which can be queried \textit{at any given time} on samples belonging to previously seen contexts $C_{\leq t}$. For simplicity, we assume that the target function is determinisic and only depends on the input, i.e. $p(Y|X,C)=p(Y|X)$. Our goal is to minimize the following objective after $T$ timesteps

\begin{align}
  \mathcal{L}_{CLP}(\theta) &\mydef \mathbb{E} \big[ \ell(f_{\theta}(X),Y) \big] \\
   &= \sum_{t=1}^T \mathbb{E}_{p(x|C_t)} \big[ \ell(f_{\theta}(X),Y) \big]
\end{align}

where $\ell : \mathcal{Y} \times \mathcal{Y} \rightarrow \mathbb{R}$ denotes a loss function between the model predictions and labels.

\paragraph{Meta CL} In Meta Continual Learning, the goal is to pretrain the predictor $f_{\theta}$ for better performance, or transfer, on the CLP objective. In the \textit{pretraining} 
 phase, we aim to learn parameters $\theta$, such that at \textit{deployment} 
time, the model can quickly adapt to any given CLP. To do so, we adopt the nested optimization procedure described in MAML \cite{finn2017model} for the pretraining step. In the inner loop, we sample a sequence of contexts $C_1, ..., C_k$ from a distribution over context sequences $p(\mathcal{C}^k)$. From this sequence, we then generate a trajectory $S_k = (X_1, Y_1), ..., (X_k, Y_k)$ with $X_i \sim P(X|C_i)$. This trajectory is equivalent to the \textit{support set} in MAML, as it mimics the non stationary learning conditions encountered at deployment time. Starting from initial parameters $\theta$, we obtain \textit{fast parameters} $\theta^k$ by processing the trajectory one (potentially batched) example at a time, performing $k$ SGD steps, with loss $\ell(f_{\theta_i}(X_i), Y_i)$ at each inner step. Denoting the one-step SGD operator with learning rate $\alpha$ as 

\begin{equation}
U \big(\theta^{0} \big) = \theta^{0} - \alpha \nabla_{\theta^{0}} \ell_0(\theta^{0}, S_{[0]}) = \theta^{1}, 
\end{equation}

\noindent
the \textit{fast parameters} $\theta_k$ are given by 

\begin{equation}
\theta^k = U \circ ... \circ U \big(\theta^0 \big).
\end{equation}

We proceed to evaluate parameters $\theta^k$ on a \textit{query set} $Q$, obtaining the \textit{outer loss} $\ell_{out}$. We recall that two key properties of a continual learner are \ca{its ability to leverage past skills to quickly learn new concepts} and to \cb{learn without negative interference on previous knowledge}. To do so, the OML~\cite{javed2019meta} objective insightfully proposes to compose $Q$ with samples from \ca{current inner loop distribution $P(X|C_{i:k})$} but also \cb{from concepts seen in previous iterations of the pretraining phase}, which we denote by $C_{old}$. The full objective can be written as 

\begin{align}
\begin{split}
\min_{\theta} &\sum_{C_{1:k} \sim p(\mathcal{C}^k)}   \text{OML}(\theta) \mydef  \\
\sum_{C_{1:k} \sim p(\mathcal{C}^k)}& \sum_{S_k \sim p(S_k|C_{1:k})}   \ell_{out} \big(U(\theta, S_k), Q  \big)
\end{split}
\end{align}

\noindent
Once $\ell_{out}$ has been evaluated, we backpropagate through the inner learning procedure, all the way to the initial $\theta$. Using this gradient we calculate a new set of initial weights $\theta \coloneqq U(\theta, \ell_{out})$ and iterate again over the whole process until convergence. 

In practice, \cite{javed2019meta} partition the full model $f_{\theta} : \mathcal{X} \rightarrow \mathcal{Y}$ into a model base (or trunk)  ${b}_{\theta_{b}} : \mathcal{X} \rightarrow \mathbb{R}^d$ which extracts a learned representation, and a model head $h_{\theta_{h}} :  \mathbb{R}^d \rightarrow  \mathcal{Y}$ which predicts the final output from the learned representation. In other words, the model can be represented as $f_{\theta} = h_{\theta_{h}} \circ b_{\theta_{b}}$. \lc{In practice, the network base typically contains all convolutional blocks, while the head is a small multi-layer perceptron.} 

During the inner loop (and deployment), only the head parameters $\theta_h$ are updated via SGD. The  parameters $\theta_{b}$ are only updated in the outer loop, meaning that latent representations for all inputs can be extracted in parallel, since $\theta_{b}$ is fixed. This modification also leads to training optimization procedure that is mode stable, leading to significantly better results.

Again, the key contribution of OML is that one can approximate the full $\mathcal{L}_{CLP}$ by artificially creating short non-stationary sequences and evaluating knowledge retention on a sample of past data. \lc{Crucially, the loss $\ell$ used in both inner and outer loop, is a supervised loss. Moreover, the non-stationary sequences are created using labelled information. In this work, we design a new algorithm for the pretraining phase which bypasses the need for labelled data.} We discuss our approach next. 

\section{Proposed Method}
In this section, we introduce the core components our method, SPeCiaL. We first discuss our unsupervised labelling procedure, enabling SPeCiaL to work on unlabelled datasets. We then discuss our mechanism to fight catastrophic forgetting. Lastly, we discuss several modifications reducing the computational and memory footprint of our method, a key requirement when scaling to larger datasets. A full workflow or our method is shown in Figure \ref{fig:cartoon}. 

\subsection{Generating Pseudo Labels}
Our labelling procedure relies on data augmentation in pixel space, in order to generate multiple transformed images, or views, that share common semantic features. Similar to instance-level classifiers, we propose to assign each image (and its generated views) to a unique label. However, unlike most instance-level classifiers, we do not require a large final linear layer, nor do we use noise contrastive estimation. Similar to OML, we reset class prototypes once their lifespan is complete, allowing for a tractable solution (more information in section \ref{DFC}). We follow the augmentation pipeline described in \cite{chen2020simple}, yielding a stochastic transformation operator consisting of three base augmentations: random cropping which is then resized back to the original size, random color distortions, and random Gaussian blur. 
We highlight that this procedure can operate online, meaning it does not require a separate isolated clustering step, as in \cite{bertugli2021generalising,hsu2018unsupervised}. Moreover, by incorporating the labelling procedure within our variant of OML, labels are not static, hence the model is less prone to overfitting, and can benefit from longer training (see section \ref{results}). 




\subsection{Maximizing parallel computations} Processing the incoming batch of data sequentially hinders the ability to run computations in parallel. We make two modifications to circumvent this issue. First, following \cite{beaulieu2020learning}, we partition the full model such that the model head contains only the last linear classification layer. Since only the head parameters have to be executed sequentially, this enables us to run the feature extraction process in parallel, which represents a significant portion of the computation.
Second, we train on \textit{multiple sequences in parallel}. \lc{During the inner loop procedure, instead of processing a single example per timestep, we instead process $M$ examples every step. This is equivalent to first generating $M$ independent streams, and processing (in batch) one example per stream at each timestep.}
This further maximizes the proportion of parellizable operations. We refer to the number of sequences ($M$) in the inner loop as the \textit{meta batch size}.

\subsection{Delayed Feature Recall} \label{DFC}The original OML objective proposes to sample across all previously seen contexts $C_{old}$ (or previously seen classes) to obtain samples for the outer loop, such that the model is incentivized to learn an update procedure which is robust to forgetting. In our current framework, this is not currently possible as it would require to keep in memory a number of vectors in the output layer that is proportional to the number of images seen. Instead, we propose to sample only from the last $N$ seen samples, by storing these recent samples in a queue. This way, only the output vectors of the samples in the queue need to be kept additionally in memory. As a consequence, the lifespan of a given label starts when an image first appears in the inner loop, and persists until the sample is fully discarded from the queue. Therefore, we only require that our model's output size be larger than the sum of the number of unique images seen within a meta-training step, and the queue size. We note that when sampling a random label for an incoming image, we make sure to exclude already assigned labels in the queue to avoid collisions.

Furthermore, instead of storing raw images in the queue, we instead store the representation produced by the model trunk $b: \mathcal{X} \rightarrow \mathbb{R}^d$. This reduces both the computational and memory footprint of our algorithm, since we are reusing previously computed features. \lc{We note that in practice, using a small queue works best. Therefore the representations are stored in the queue for a few iterations only, and do not become stale, so no additional mechanism to counter staleness \cite{he2020momentum} is needed}. The full workflow of our method can be seen in Figure \ref{fig:cartoon}.


\section{Experiments}
\label{experiments}

We design our experiments to evaluate how well the learned representations can transfer when learning new classes from a common domain.

\begin{figure}[t]
    \centering
    \includegraphics[width=.25\textwidth]{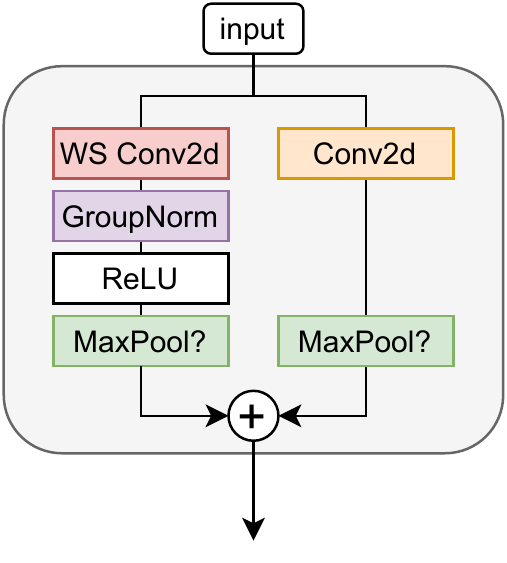}
    \caption{Architecture Block. WS stands for Weight Standardized Convolutions. This block is used in the network base, denoted in red in Figure \ref{fig:cartoon}.}
    \label{fig:arch_block}
\end{figure}

\begin{figure*}
    \begin{minipage}{0.95\textwidth}
    \includegraphics[width=.5\textwidth]{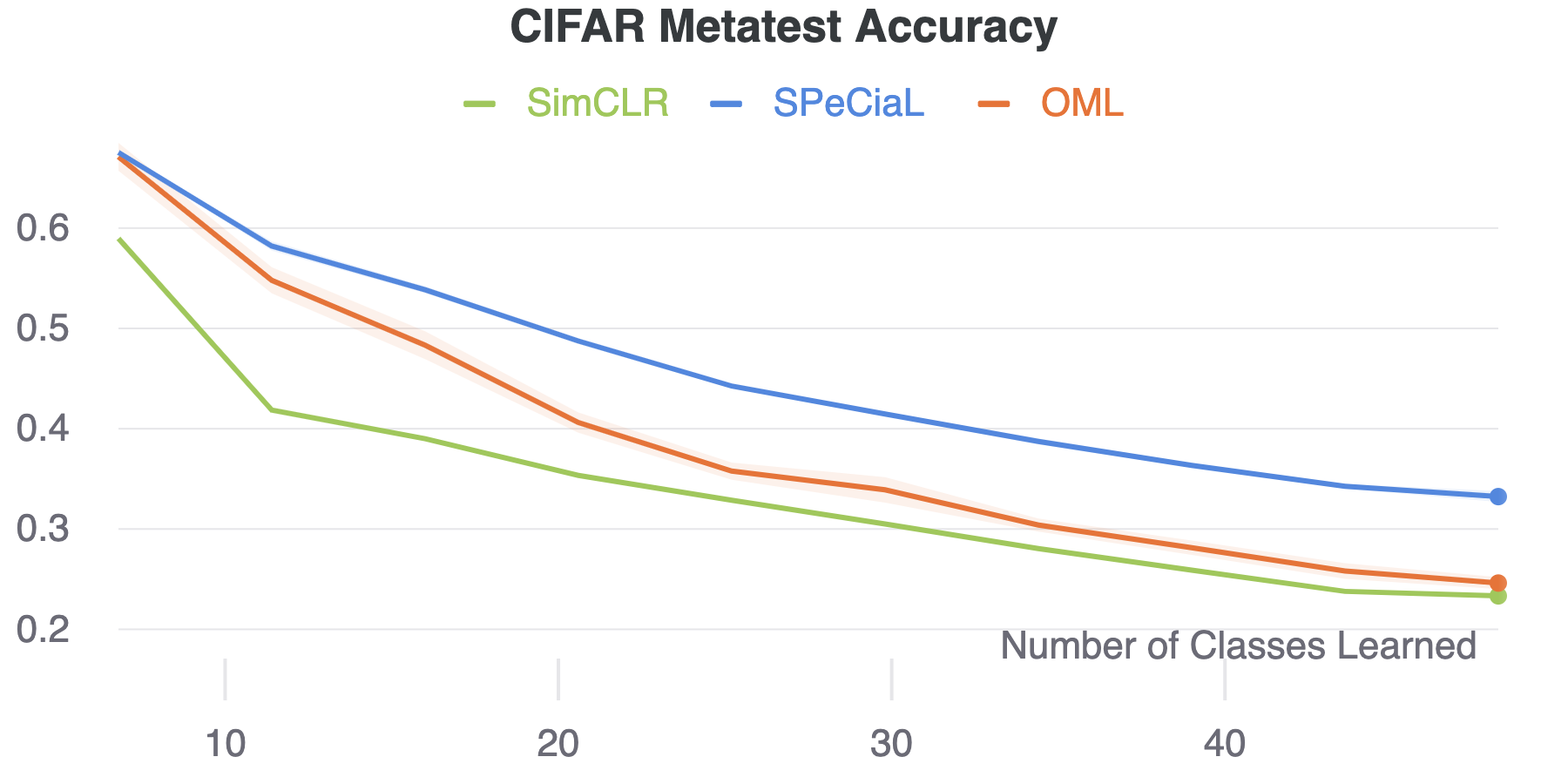}
    \includegraphics[width=.5\textwidth]{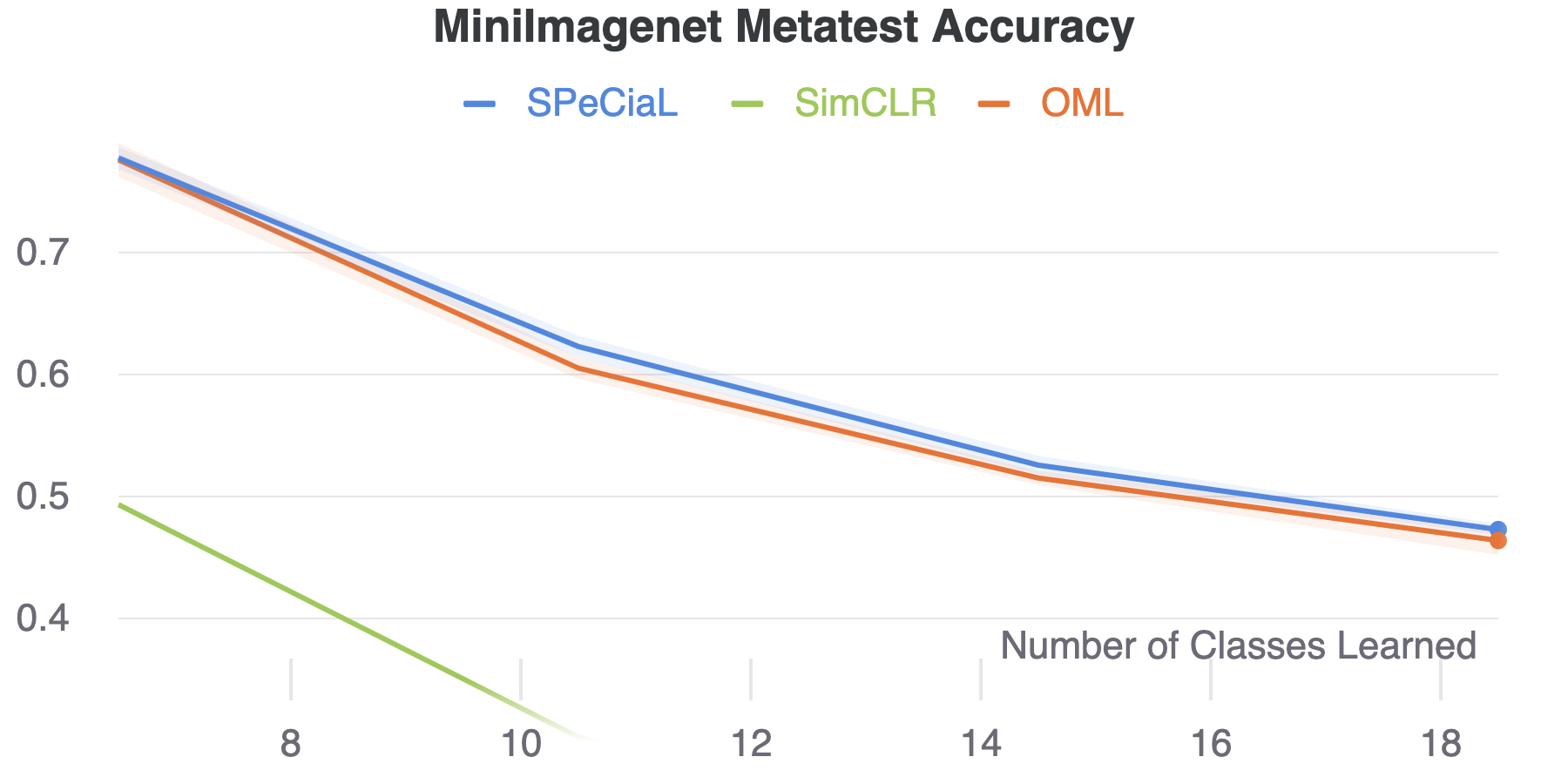}
    \end{minipage}\hfill
    \vspace{10pt}
    \caption{MetaTest Accuracy during training of different models. Curves are averaged over three runs, and standard error is shaded. Smoothing is applied for clarity.}
    \label{fig:results}
\end{figure*}

\paragraph{Pretraining} \lc{In practice, for each meta-train step we instantiate a Continual Learning Problem as follows: we first sample a set of (e.g. 5) context variables $\{C_1, .., C_5\}$. From this set, we create a sequence of correlated contexts by repeating each context $n$ times : $(C_1^{(1)}, \hdots , C_1^{(n)}, C_2^{(1)} \hdots, C_5^{(n)})$ 
. In the supervised case, we create a context $C_i$ for each distinct label $Y_i$ in the meta-training set. Therefore, sampling from $P(X_i|C_i)$ is equivalent to sampling from the class-conditional distribution $P(X_i|Y_i)$. For the unsupervised case, we create a context $C_j$ for each image in the dataset, and sampling from $P(X_j|C_j)$ yields different augmented views of the $j$-th image in the meta-training set. For all methods considered, we cross validate the optimal value for the meta batch size.}

\paragraph{Deployment: Few Shot Continual Learning}
Following \cite{javed2019meta,beaulieu2020learning,bertugli2021generalising}, our instantiation of a CLP follows the continual few shot learning procedure. At metatest time, we fix all layers except the last one (i.e. the RLN layers), as done in the inner loop procedure. We then proceed to learn a linear classifier (PLN) from a stream of $C$ classes, with $N_{c}$ samples for each class. The data is streamed example by example, one class at a time. In other words, the learner does $N_c \times C$ updates, seeing $N_c \times C$ samples (once), and observing $N_c - 1$ distribution shifts.

\paragraph{Datasets}
We experiment on two datasets. We first employ the MiniImagenet \cite{vinyals2016matching} dataset, commonly used in few shot leaning. It comprises 100 classes, with 64 used for meta-training, 16 for meta-validation and 20 for meta-testing. Images are resized to $84 \times 84$. We also use the CIFAR-10/100\cite{krizhevsky2009learning} datasets. All the meta-training is performed on CIFAR-10, and we do an 50-50 split for CIFAR100 for meta-validation and meta-test respectively. Images are resized to $32 \times 32$.

\paragraph{Architecture}
For all baselines considered, we use a 4 block residual network. Each block consists of two computation paths, each with a single convolution. We use Weight Standardized \cite{qiao2019weight} convolutions with Group Normalization \cite{wu2018group}, a normalization approach with does not use batch-level statistics and has shown strong performance in small batch settings. We use MaxPooling to downsample our input when required, i.e. in the all blocks for MiniImagenet, and the first three blocks for CIFAR. The block's architecture is illustrated in Figure \ref{fig:arch_block}. The key difference with previous architectures commonly used in  Meta Learning, is the use of a computational path with no activations. We further discuss the impact of these architectural changes in section \ref{arch_disc}.

\paragraph{Baselines}
We mainly compare against OML, where supervised pretraining is used. Our OML implementation can optionally leverage tools specific to the SSL pipeline, namely the use of augmentations. We also benchmark against  SimCLR \cite{chen2020simple}, a strong and flexible SSL model. For all approaches, we use the same architecture and training budget, allowing us to compare methods on equal footing and better isolate the contribution of our method. 

\paragraph{Pretraining}
Models are trained for 350,000 optimization steps. For OML and SPeCiaL, a step corresponds to a full inner / outer loop, while for regular models it accounts for the processing of one minibatch. We found that for meta models, the AdamW optimizer with gradient clipping yielded best results. For other models, we used SGD with momentum. All models use a learning rate scheduler consisting of a linear warmup with cosine decay. For OML and SPeCiaL, we used cross-validation to determine the optimal sequence length, meta batch size and the number of distinct images per sequence. For SimCLR, we use the largest batch size which fits in memory, and the (batch size) adaptive learning rate procedure described in the paper. All experiments are executed on one Tesla V100 GPU.

\paragraph{MetaTesting}
Once the models are trained, we evaluate their performance on the downstream Continual Few Shot Learning Task. As in \cite{javed2019meta,bertugli2021generalising} we set the number of examples per class to 30. We follow \cite{javed2019meta} and cross-validate the learning rate for each metatest trajectory length reported. 


\begin{figure*}[t]
    \begin{minipage}{0.95\textwidth}
    \includegraphics[width=.5\textwidth]{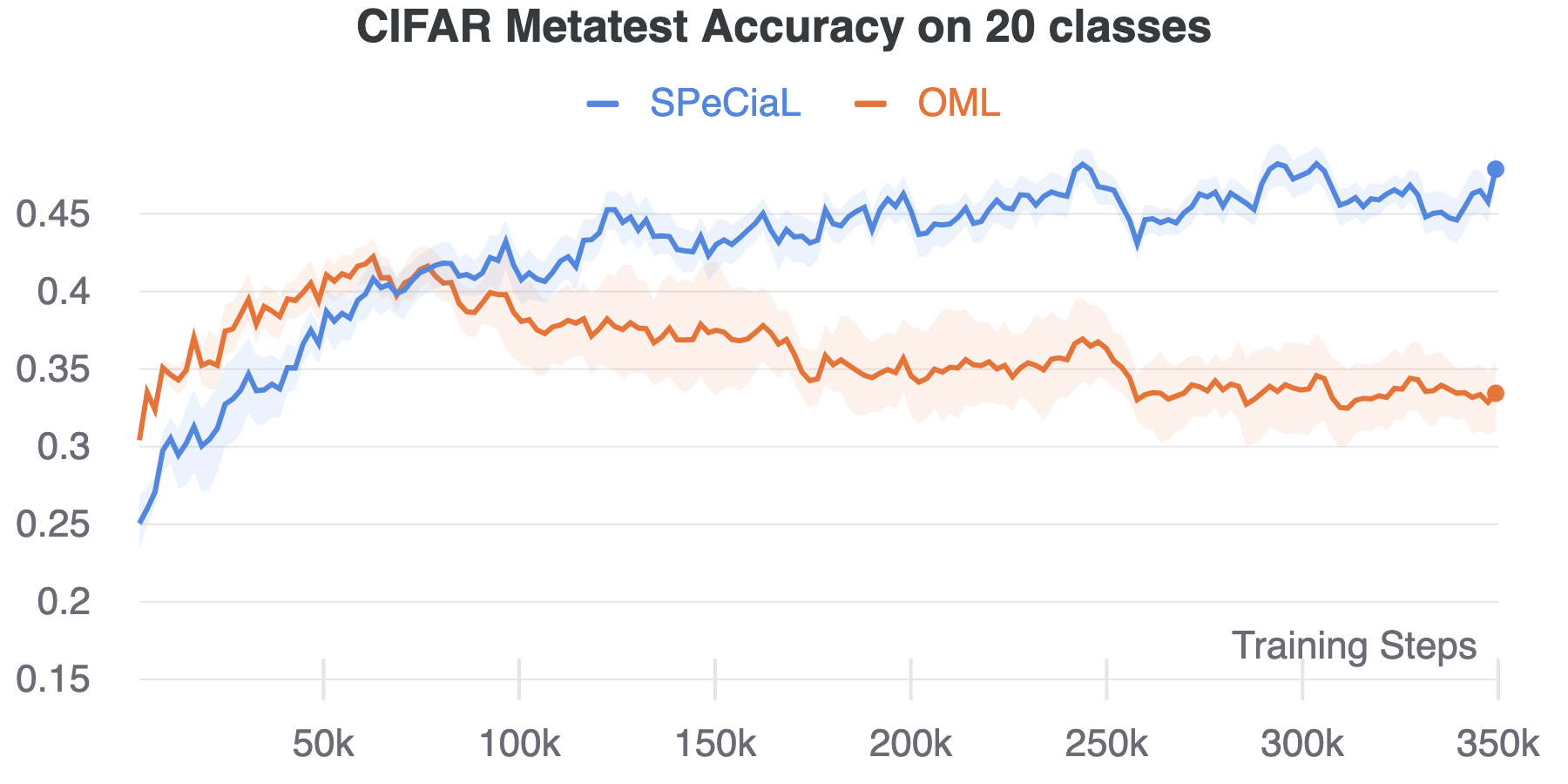}
    \includegraphics[width=.5\textwidth]{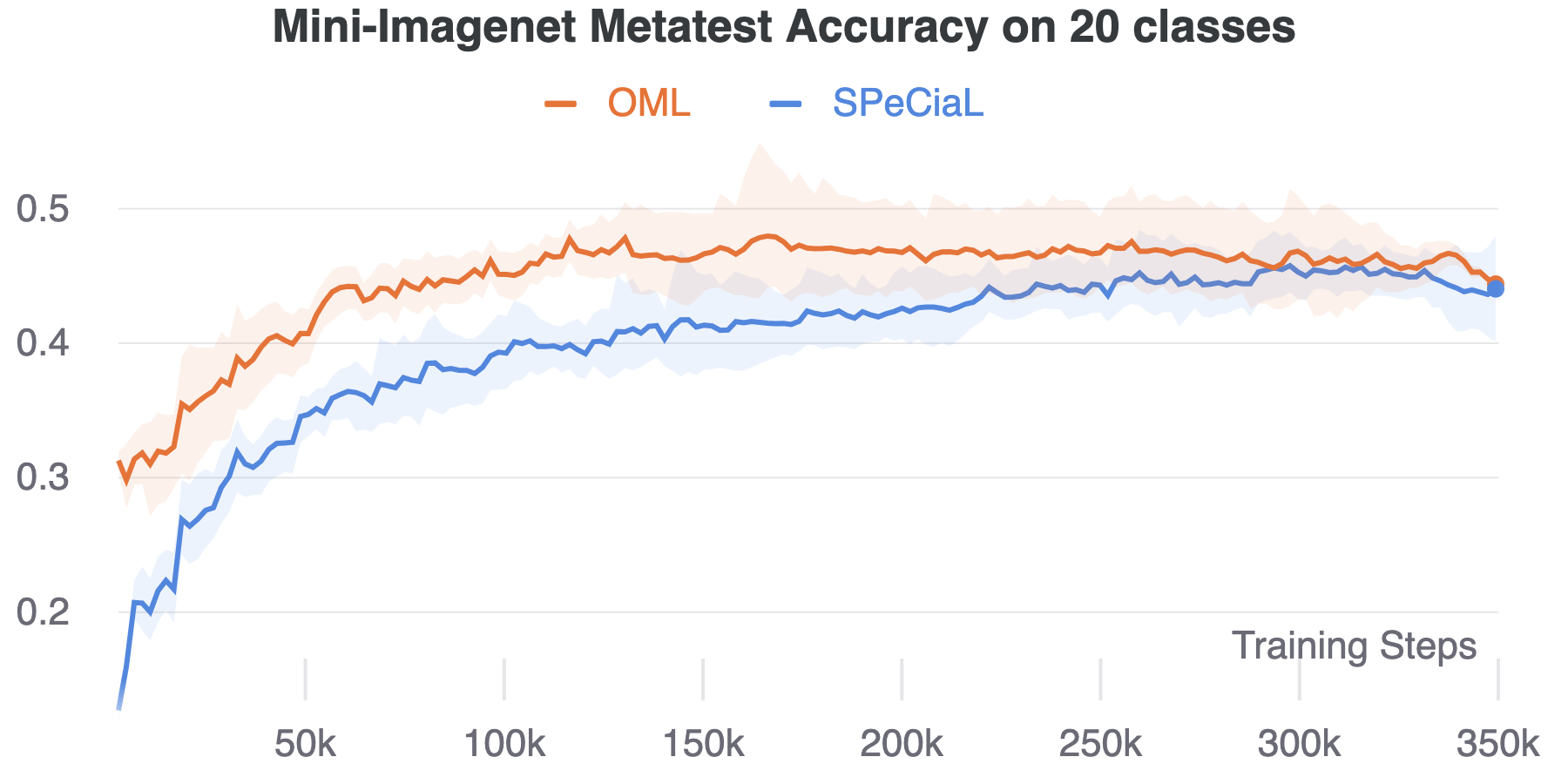}
    \end{minipage}\hfill
    \vspace{10pt}
    \caption{MetaTest Accuracy during training of different models. Curves are averaged over three runs, and standard error is shaded. Smoothing is applied for clarity.}
    \label{fig:curves}
\end{figure*}

\subsection{Results} \label{results}
We first present results for the CIFAR experiment in Fig \ref{fig:results}. In this experiment, SPeCiaL is actually able to outperform its fully supervised counterpart. Moreover, this performance gap widens as more tasks are being learned, suggesting that the representations learned by our model are better adapted for longer (and more challenging) scenarios. We generally found that self supervised method, e.g. SimcLR and SPeCiaL, can greatly benefit from longer training. Unlike OML, which starts overfitting after 100K steps, SPeCiaL continues to improve when given more computation. We illustrate this in Figure \ref{fig:curves}. This observation is consistent with \cite{assran2020supervision}, who note that when combining supervised and SSL losses, after a fixed number of epochs the supervised signal needs to be removed to avoid a decrease in performance.  

We proceed with the MiniImagenet experiment, shown on the right of Fig \ref{fig:results}. Here SPeCiaL is able to match the performance of OML. We note here that OML does not overfit as on the smaller scale CIFAR dataset. Instead, performance stagnates after 150K steps. 

To summarize, early in training OML outperforms other unsupervised methods. As training progresses, SpeCiaL is able to close the performance gap, and benefits more from longer training. 

\begin{figure}[h]
    \centering
    \includegraphics[width=.45\textwidth]{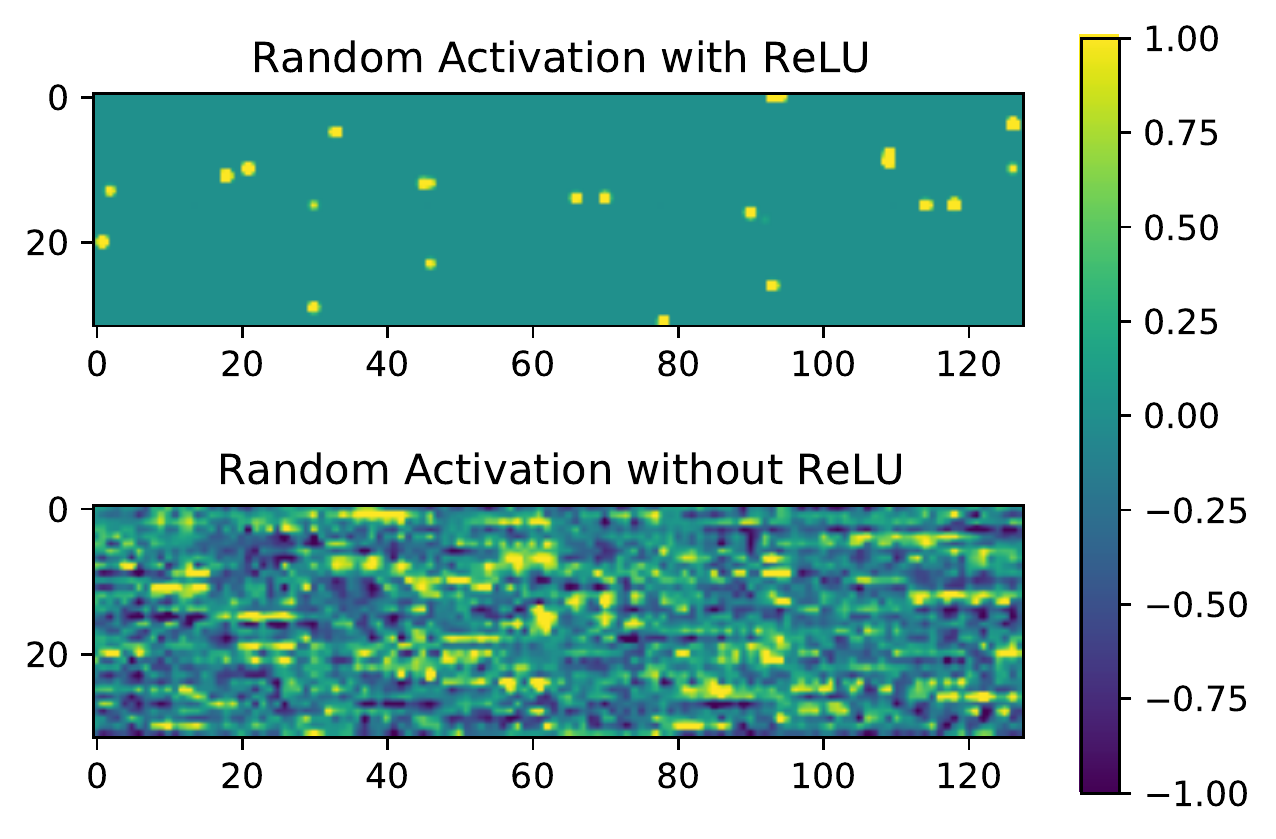}
    \caption{Impact of ReLU activations on the sparsity of the learned representations. Bottom image shows a random representation obtained by our method, SPeCiaL, with the architecture presented above. Top picture is a random representation when we add a ReLU activation after the RLN output.}
    \label{fig:activations}
\end{figure}

\subsection{Sparse Representation is Architecture-Dependent}

\label{arch_disc}
In previous Meta Continual Learning paper, it was shown that the model learns sparse representations as a way to mitigate catastrophic forgetting. The idea of using sparse representations has been proposed in earlier work \cite{french1991using}, and revisited in modern CL methods \cite{liu2019utility,aljundi2018selfless}. Therefore, seeing sparsity emerge in a data-driven way from MCL algorithms, \lc{as reported in \cite{javed2019meta,beaulieu2020learning}}, confirms that it is a good inductive bias for continual learning. In our work however, we found that the OML objective alone is not sufficient to obtain sparse representations : one must also use a suitable architecture for sparsity to naturally emerge. In the original work, the architecture used consisted of Convolutions and ReLU activations, hence the last operation performed by the RLN was a nonlinearity favorable to cancelling neurons. 

We run an experiment where we add an additional ReLU activation after the final block of the RLN. We compare the activations obtained with this variant in Figure \ref{fig:activations}. We see that indeed sparsity emerges, which is not the case under the base architecture. Moreover, in the base architecture used in our experiments (see Fig. \ref{fig:arch_block}), one computational path finishes with a ReLU + MaxPool operation. Therefore the model could in practice only leverage this path and obtain sparse representations, without the explicit ReLU after the skip connection. Yet, we found that this skip connection is leveraged by the model, and overall stabilized training. We therefore conclude that the architecture used plays an important role in shaping the activation patterns. Moreover, we found that this additional ReLU activation after the final skip connection gave a model which was harder to optimize, and therefore led to worse performance. 

\section{Conclusion} 
In this work we proposed a novel unsupervised meta objective to learn representations suitable for Continual Learning. We showed that when deployed in a Continual Learning setting, the learned representations can match or exceed the performance of fully supervised meta algorithms. Furthermore, we showed that sparsity does not always occur when using Meta Continual Learning algorithms, and that the architecture of the model plays a significant role in this context. 

Our work takes a first step towards leveraging unlabelled data to prepare a system before its deployment on non-stationary data. An important next step for this work is to investigate how we can scale up meta algorithms presented in this work, and see if they can excel in the large data regime like Self-Supervised Learning methods.

\newpage
\bibliographystyle{named}
\bibliography{ijcai21}

\end{document}